\documentclass[letterpaper, 10 pt, conference]{ieeeconf}  
\IEEEoverridecommandlockouts
\usepackage{cite}
\usepackage{amsmath,amssymb,amsfonts}
\usepackage{algorithmic}
\usepackage{graphicx}
\usepackage{textcomp}
\usepackage{xcolor}
\def\BibTeX{{\rm B\kern-.05em{\sc i\kern-.025em b}\kern-.08em
    T\kern-.1667em\lower.7ex\hbox{E}\kern-.125emX}}
\usepackage{multirow}
\usepackage{subcaption}
\usepackage{caption} 
\PassOptionsToPackage{hyphens}{url}\usepackage{hyperref}
\usepackage{hypcap}
\usepackage{tcolorbox}

\usepackage{comment}
\usepackage{comment}
\usepackage{xcolor, color, soul}
\sethlcolor{yellow}
\usepackage{gensymb}
\usepackage{lipsum}
\usepackage{graphicx}
\usepackage{booktabs}
\usepackage{balance}

\renewcommand{\baselinestretch}{0.98}

\title{\LARGE \bf
Gaze-supported
Large Language Model Framework for
Bi-directional Human-Robot Interaction}

\author{Jens V. Rüppel$^{1}$,
       Andrey Rudenko$^{2}$,
       Tim Schreiter$^1$,
       Martin Magnusson$^3$,
       and~Achim J.~Lilienthal$^{1,3}$
\thanks{$^{1}$Technical University of Munich, Germany  {\tt\small \{jens.v.rueppel,tim.schreiter, achim.j.lilienthal\} @tum.de}}%
\thanks{$^{2}$Robert Bosch GmbH, Corporate Research, Stuttgart, Germany}
\thanks{$^{3}$Centre for Applied Autonomous Sensor Systems (AASS), \"Orebro University, Sweden} 
\thanks{This work was supported by the European Union’s Horizon 2020 research and innovation program under grant agreement No. 101017274 (DARKO).}
}

\begin{document}
\maketitle



\begin{abstract}
The rapid development of Large Language Models (LLMs) creates an exciting potential for flexible, general knowledge-driven Human-Robot Interaction (HRI) systems for assistive robots. Existing HRI systems demonstrate great progress in interpreting and following user instructions, action generation, and robot task solving. On the other hand, bi-directional, multimodal, and context-aware support of the user in collaborative tasks still remains an open challenge. In this paper, we present a gaze- and speech-informed interface to the assistive robot, which is able to perceive the working environment from multiple vision inputs and support the dynamic user in their tasks. Our system is designed to be modular and transferable to adapt to diverse tasks and robots, and it is real-time capable due to the language-based interaction state representation and fast on-board perception modules. Its development was supported by multiple public dissemination events, contributing important considerations for improved robustness and user experience. Furthermore, in a lab study, we compare the performance and user ratings of our system with those of a traditional scripted HRI pipeline. Our findings indicate that an LLM-based approach enhances adaptability and marginally improves user engagement and task execution metrics but may produce redundant output, while a scripted pipeline is well suited for more straightforward tasks.
\end{abstract}

\section{INTRODUCTION}\label{sec:intro}





Research in human-robot interaction (HRI) in workplace settings aims to develop methods that enable robots to effectively collaborate with human workers and support diverse and dynamic tasks. Although recent frameworks have shown promising results in task planning \cite{zhi2024closed}, robot control \cite{wang2024lami}, and the execution of user commands \cite{ali2024comparing}, they often cannot provide intelligent, adaptive assistance that can respond flexibly to human actions in real-time, particularly in open-ended and unstructured environments.

Modern HRI frameworks often lack dynamic adaptability in response to the changes in the environment that occur as users perform tasks independently or collaboratively with robots. Scripted procedures in these interactions often anticipate specific user input to maintain interaction flow and may delay or fail completely if the environment reaches an unpredictable state \cite{schreiter2025evaluating}. Furthermore, these frameworks are often designed and optimized for a specific task \cite{huang2016anticipatory}, executed by a specific robot, lacking the ability to adapt quickly to different scenarios.
In recent years, the research community has progressed towards the general embodied AI, transcending the above-mentioned issues with advanced environment perception and reasoning capabilities.

\begin{figure}[t]
\centering
\includegraphics[width=0.49\columnwidth]{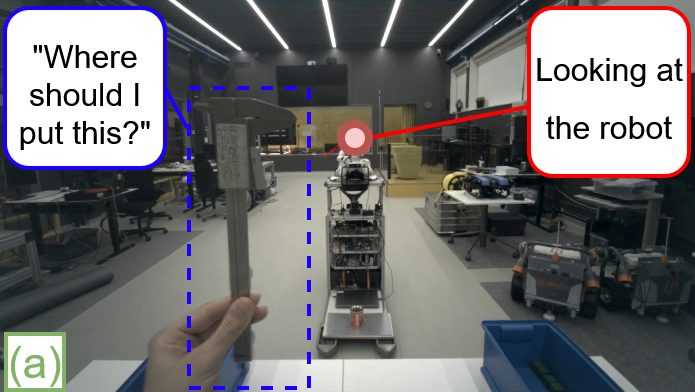}
\includegraphics[width=0.49\columnwidth]{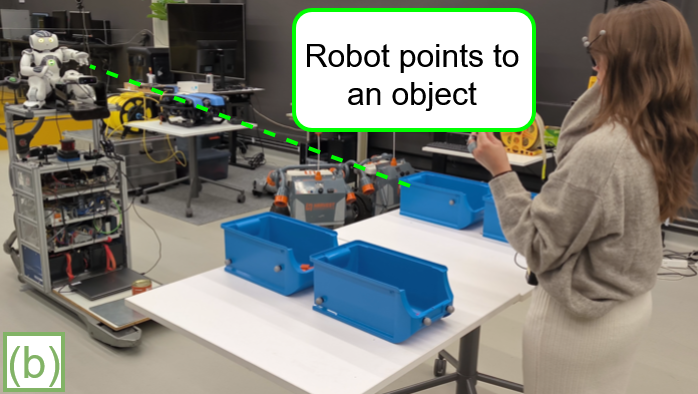}\\
\vspace{3pt}
\includegraphics[width=0.49\columnwidth]{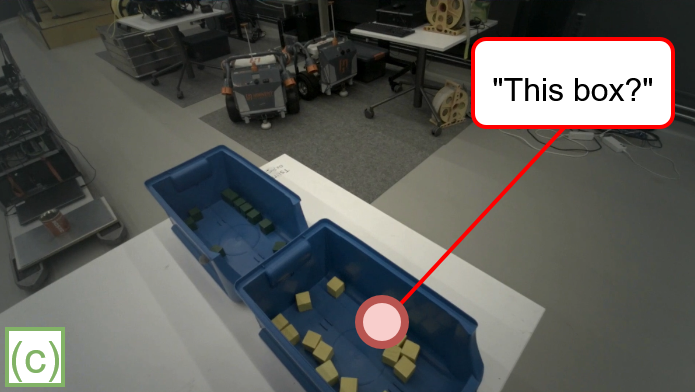}
\vspace{-3pt}
\includegraphics[width=0.49\columnwidth]{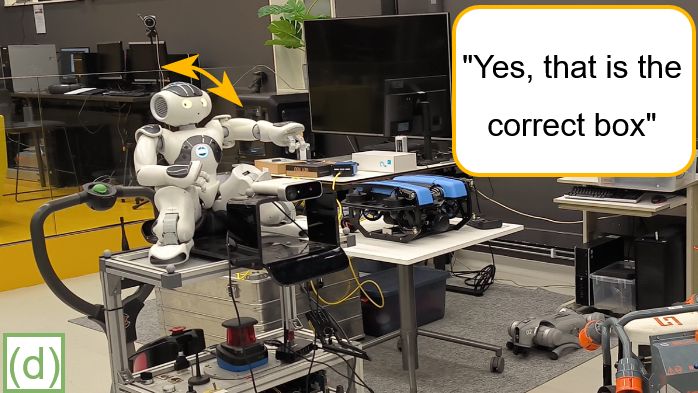}
\caption{Bi-directional gaze- and speech-supported communication between the robot and the user, achieved by our framework:
\textbf{(a)} the user is asking the robot where to put the object,
\textbf{(b)} the robot is pointing to the box, \textbf{(c)} the person is looking at one of the boxes on the table, asking for confirmation, \textbf{(d)} the robot makes a nodding gesture and vocally confirms.
}
\label{fig:cover}
\end{figure}

In this paper, we present a novel user-centric HRI framework with multimodal in- and outputs, and distributed perception of the 3D environment. The framework uses a Large Language Model (LLM) based backbone to ground the inputs and provide task decomposition from high-level verbal descriptions to low-level robot actions. Specifically, we use cues of the environment and the user, measured onboard with robot sensing and user gaze tracking, accompanied by a speech interface. User input is processed using the LLM backbone, and structured plans are provided using Chain-of-Thought (CoT) reasoning. This approach can detect the user's intention or confusion and produce context-appropriate responses and actions, see an example in Figure \ref{fig:cover}. 

Our work also contributes towards transferable robotic intent communication. The framework can be fitted to different robot platforms with minimal effort, reducing the need for task-specific modifications or reprogramming. To demonstrate this, we evaluate our approach using the ``Anthropomorphic Robot Mock Driver'' (ARMoD) robot communication concept for non-humanoid robots
\cite{schreiter2022effect, schreiter2023advantages, schreiter2024thor}.

%



The development of our system was supported by several public dissemination events, in which we could investigate the real-time real-world performance of the various perception and reasoning components and increase the robustness of the developed solutions. 
For the validation of our framework, we conducted two multi-step spatial interaction experiments (parts of which are already discussed in prior work \cite{schreiter2025evaluating}), where participants had to engage in collaborative tasks with a mobile robot that communicated via an ARMoD. The aim of our study was to investigate whether our transferable, multimodal HRI framework outperforms an equivalent static, pre-defined interaction in a manufacturing HRI context. We found that both configurations perform on par in terms of task efficiency and engagement in well-defined, streamlined tasks, with the LLM condition receiving marginally higher responsiveness ratings.

\begin{figure}[t!]
\centering
\includegraphics[width=0.8\columnwidth]{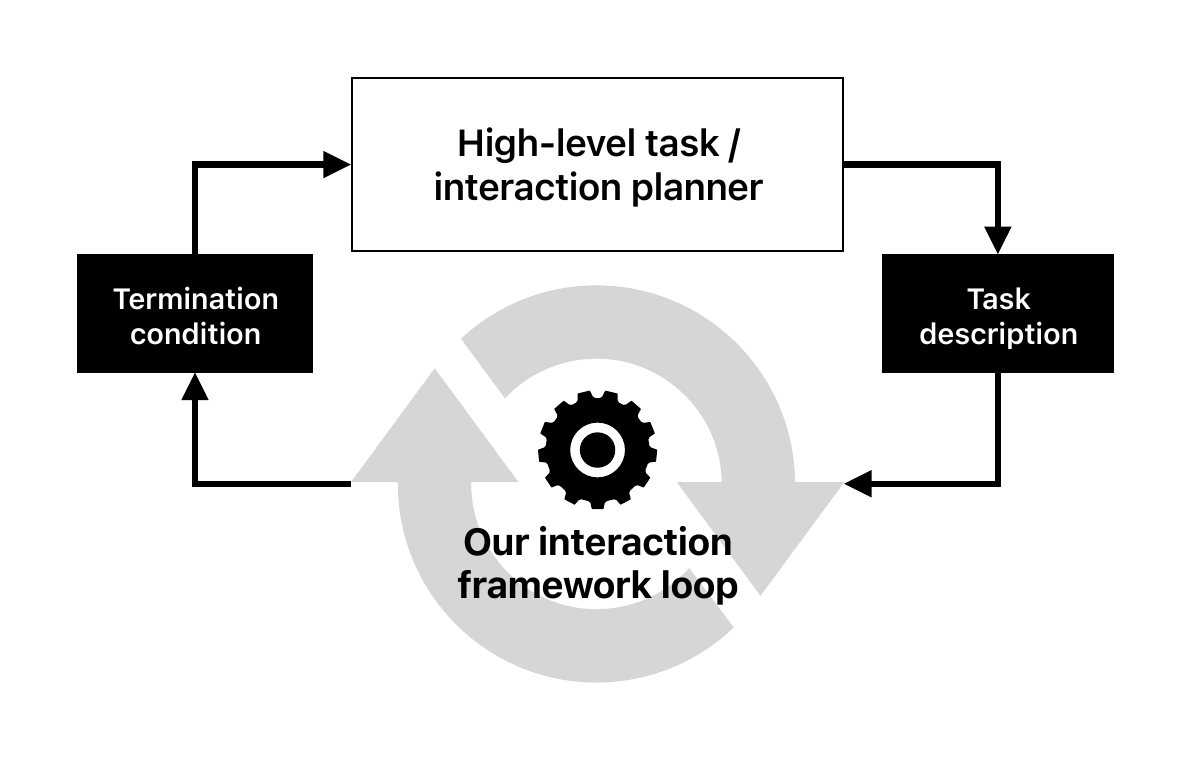}
\vspace{-0.1cm}
\caption{Our framework operates as a task-oriented interaction loop, triggered by a high-level HRI module or task planner.}
\label{fig:interaction_loop}
\end{figure}

\section{RELATED WORK}





Recent developments in AI systems led to the rise of LLMs with strong context-aware generalization and reasoning capabilities, offering a promising solution to many HRI use cases \cite{atuhurra2024leveraging}. While many works address tasks like navigation \cite{zhi2024closed} and planning \cite{vemprala2024chatgpt, zhi2024closed}, systems that focus on multimodal communication and user activity support are still scarce. Wang et al. \cite{wang2024lami} develop a multimodal HRI approach that takes several inputs, including gaze and voice, and generates multimodal actions based on the analysis of the current situation. They define atomic actions to generate a variety of different robot behaviors. 
Researchers incorporated multimodality in an HRI framework before, by focusing on object categorization and group discussions \cite{tanneberg2024help}. However, the spatial environment is very limited to interactions on a table, and humans cannot freely walk around.

When working on shared tasks, communication is important to understand each other's goals, intentions, and actions. It fosters collaboration by making the intentions of both humans and robots visible \cite{pascher2023communicate} and facilitates real-time feedback and action corrections. Furthermore, it makes human-robot collaboration safer in shared spaces \cite{chadalavada2020bi}.
Natural language is a common form of verbal communication in modern HRI applications \cite{allgeuer2024robots, wang2024lami, han2021need}.
Communicating with robots in spoken language is a fast and efficient approach, allowing, for example, real-time coordination and feedback to physical actions, which can enhance user experience 
\cite{marge2022spoken}. It can also help robots communicate their intent by verbally expressing their next moves to meet the user's expectations.
\cite{han2021need}.



In addition to bi-directional application of speech, gaze tracking facilitates 
understanding of the user's intention in the surrounding 3D environment, especially if the tasks require moving in space \cite{schreiter2024human}. Gaze-support facilitates joint attention by narrowing the possible three-dimensional space to specific points of interest \cite{moon2014meet, iwasaki2024perceptive}, and conveys important non-verbal interaction feedback. In HRI applications, gaze patterns can indicate the need for assistance in complex tasks \cite{kurylo2019using}. Examples for gaze tracking in HRI tasks are \cite{wang2024lami, huang2016anticipatory}. However, they tend to be restricted to a single robot or specific task and often do not generalize well. 


In contrast to that, our framework includes multimodal input and output with speech, gaze, and 3D environment perception, to improve generalization and transferability to other robots and tasks through a modular design and supported by LLM reasoning. Reasoning expands the functionality of LLM-supported systems toward more real-time decision making in open-ended environments with unpredictable states \cite{alijoyo2024enhancing}, supports situational awareness, and enables contextual reasoning about tasks and environmental states
\cite{alami2006toward}. By leveraging reasoning techniques, ambiguous scenarios can be clarified, and adaptive strategies for overcoming undefined states can be devised.

Recent works combined gaze and language for HRI applications using LLMs. Lai et al. \cite{lai2025fam} developed a system constrained by specific hardware configurations and predefined tasks, and Menendez et al. \cite{menendez2025semanticscanpath} focused on tabletop environments, relying on marker-based object detection and head orientation as a proxy for gaze direction. However, neither study conducted user evaluations to assess their systems' impact on human-robot interaction. Our approach addresses these limitations by implementing open-world object detection combined with high-precision mobile eye tracking and unrestricted user movement. Furthermore, we validate our system through controlled user studies that compare our multimodal approach against a scripted baseline.


In summary, previous systems were often designed for a spatially limited and narrow environment using predefined actions and tasks. Their use is often restricted to a single robot with a limited set of actions \cite{wang2024lami, tanneberg2024help, ali2024comparing, lai2025fam}. In this paper, we present a novel approach to combine the functionality of a multimodal HRI interface with a comprehensive reasoning module viable to be applied to a wide variety of different robots with minimal implementation efforts.

\begin{figure*}[t!]
\vspace{0.2cm}
\centering
\includegraphics[width=2\columnwidth]{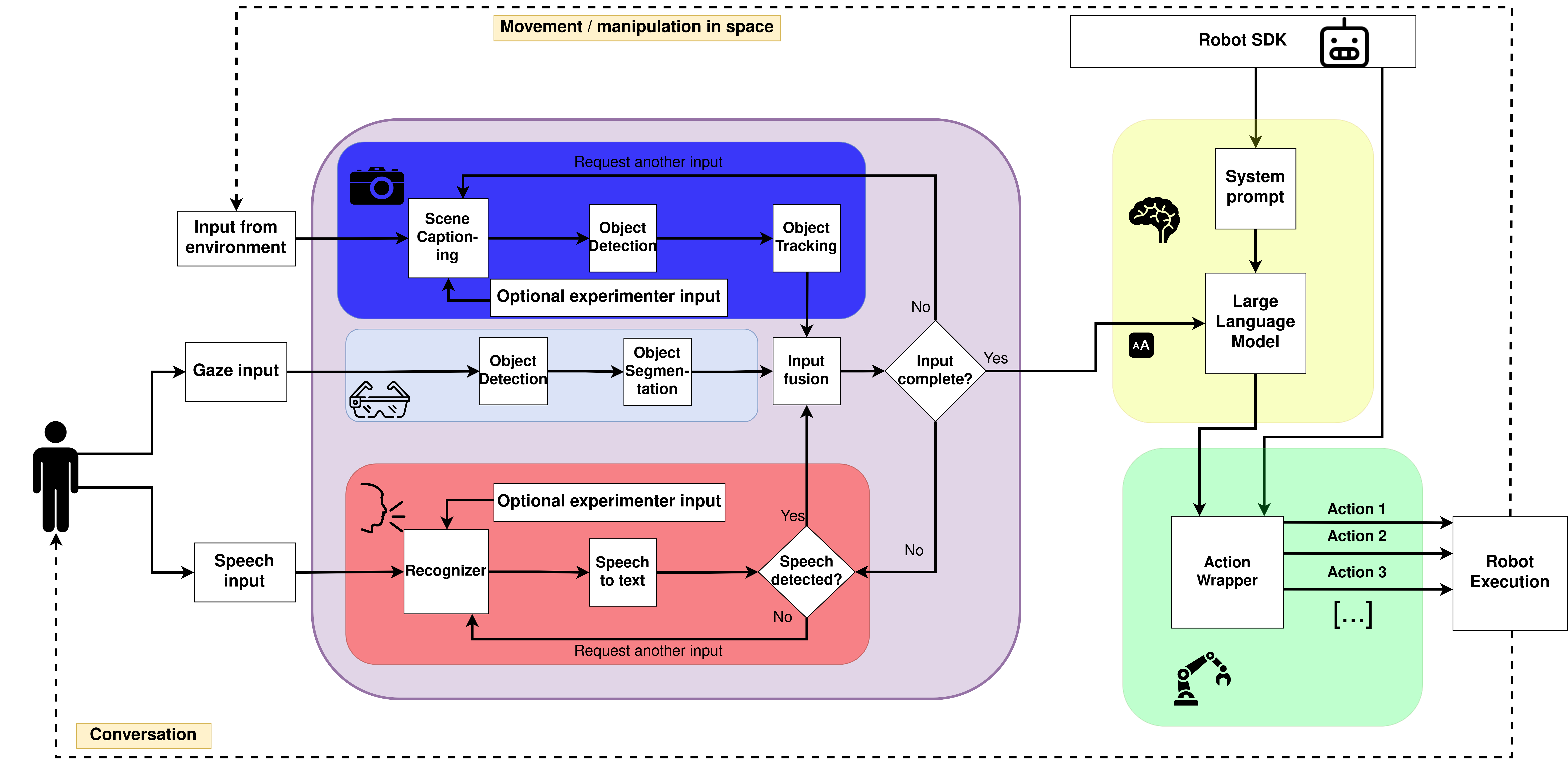}
\caption{Overview of the framework architecture. The system includes an input fusion module (purple) that integrates audio (red), video (dark blue), and mobile eye-tracker (light blue) data. The fused inputs are then passed to a reasoning module (yellow), which generates high-level task instructions for the action module (green). These modules incorporate details from the high-level task planner and the robot SDK to define necessary functionalities. Input fusion, speech, gaze, environmental input, and optional experimenter input is continuously processed.}
\label{fig:framework}
\end{figure*}

\section{DEVELOPED FRAMEWORK}

Our framework is designed as an HRI task loop, controlled by a high-level task planner or HRI component of the robot, as shown in Figure~\ref{fig:interaction_loop}. As the task request is being generated by the task planner, our system updates 
the request prompt, 
generates robot actions and maintains the interaction loop until the termination condition is reached, e.g the request is successfully resolved. All components of the framework run locally on a Dell Precision 7680 Workstation equipped with an NVIDIA RTX 3500 Ada (12GB VRAM) GPU and 64GB of RAM. Communication between the workstation and robots was handled via ROS. The interaction loop is presented in Figure~\ref{fig:framework} and detailed in Section~\ref{sec:framework_overview}, following the problem definition in Section~\ref{sec:problem_formulation}.

\subsection{Problem Formulation}
\label{sec:problem_formulation}



As the specific use-case for our framework, we consider application scenarios where the robot and the human should communicate hands-free while moving in space and manipulating objects. It could, for instance, be a mobile hospital robot supporting the patient, or a collaborative manufacturing robot intuitively understanding the next tasks from the user's instructions, supported by their gaze. On the other hand, our system can support a user operating in an unknown environment, aided by the AI reasoning and distributed vision setup, for instance, in montage, inspection, and monitoring applications. Purely vocal instructions are ambiguous and/or tedious in these cases, whereas gaze naturally 
connects vocal instructions to the dynamic 3D environment.

\subsection{Framework Overview}
\label{sec:framework_overview}

In order to make our solution transferable to diverse tasks, robots, and sensor configurations, we aim for a modular and expandable design.
Our framework runs locally to support applications with limited internet access, for instance, in
crowded environments, large industrial facilities, construction, or extraction robot applications. 

Our framework combines several modules for perception and reasoning, as shown in Figure \ref{fig:framework}: with vision (dark blue), gaze (light blue) and audio (red) input, multiple modalities are covered using state-of-the-art computer vision and speech-to-text processing techniques. The combined input is queried into the LLM-based reasoning module (yellow), producing expressive robot behavior to be executed by the action module (green). 


\subsubsection{Input Fusion Module}




The input fusion module gathers the vision and audio inputs from multiple sources and combines them into one entry. The module includes categories and locations of objects, a high-level semantic description of the scene, and transcribed voice input. The data gets collected, processed, and condensed into a Python dictionary, serving as input for further reasoning processes. The system is fully capable of running automatically without the supervision of an experimenter. In addition to that, we add intervention possibilities to manipulate its behavior, which is motivated by our dissemination findings in Section~\ref{sec:dissemination}.

Visual input is essential to provide a comprehensive representation of both the user's and the robot's perspectives within the scene. The integration involves multiple RGB(\mbox{-}D) camera feeds, capturing the required viewpoints from the robot and the user, and facilitating the bi-directional communication approach. In order to provide a context-rich understanding of the scene, the visual input undergoes several processing steps. Firstly, a BLIP-v2 model \cite{li2022blip} provides a high-level scene caption, creating the robot's semantic understanding of the scene. For object detection, the framework uses YOLO-World as a lightweight solution \cite{cheng2024yolo}. Additionally, the module uses object segmentation via a Segment Anything 2 model to make it robust to object occlusions \cite{ravi2024sam}. Following these steps, the system obtains the object categories and their location in world coordinates, grounding the robot in its environment.

Aligned with an expandable design principle, it is possible to add additional video sources and integrate them into the input fusion module to expand the visual scene representation, e.g., to cover larger areas such as multiple locations in a warehouse or an airport. To provide the LLM with information about the user's attention state, we utilize video and gaze input from a mobile eye tracker (Tobii Glasses 3), which continuously captures the user's perspective. The video and gaze data are processed through an object detection and segmentation pipeline, mapping the user's gaze direction to objects in the scene. 

The voice transcriptions are covered by a real-time transcription algorithm and a Whisper speech-to-text module \cite{radford2023robust}. It transcribes word by word and concludes them into phrases. If it stops detecting speech after the user said something, the algorithm waits 3 seconds in order to account for a short pause. 
Afterward, the phrase is considered complete and sent to the input fusion procedure only once to prevent the processing of continuous input. 
The system is designed to cover a wide range of audio sources, e.g., Bluetooth microphones or devices in the network, by capturing data from an RTSP stream. A custom solution was developed to capture microphone data from the mobile eyetracking glasses to extend their original Python API functionality, which does not support streaming audio from the microphone.

The registration of speech triggers the subsequent processing steps. The system only fuses its input and creates the Python dictionary if speech is registered. The information in the Python dictionary includes the vision inputs from the moment the transcription is done.




\begin{figure}[t!]
\vspace{0.2cm}
\centering
\includegraphics[width=\columnwidth]{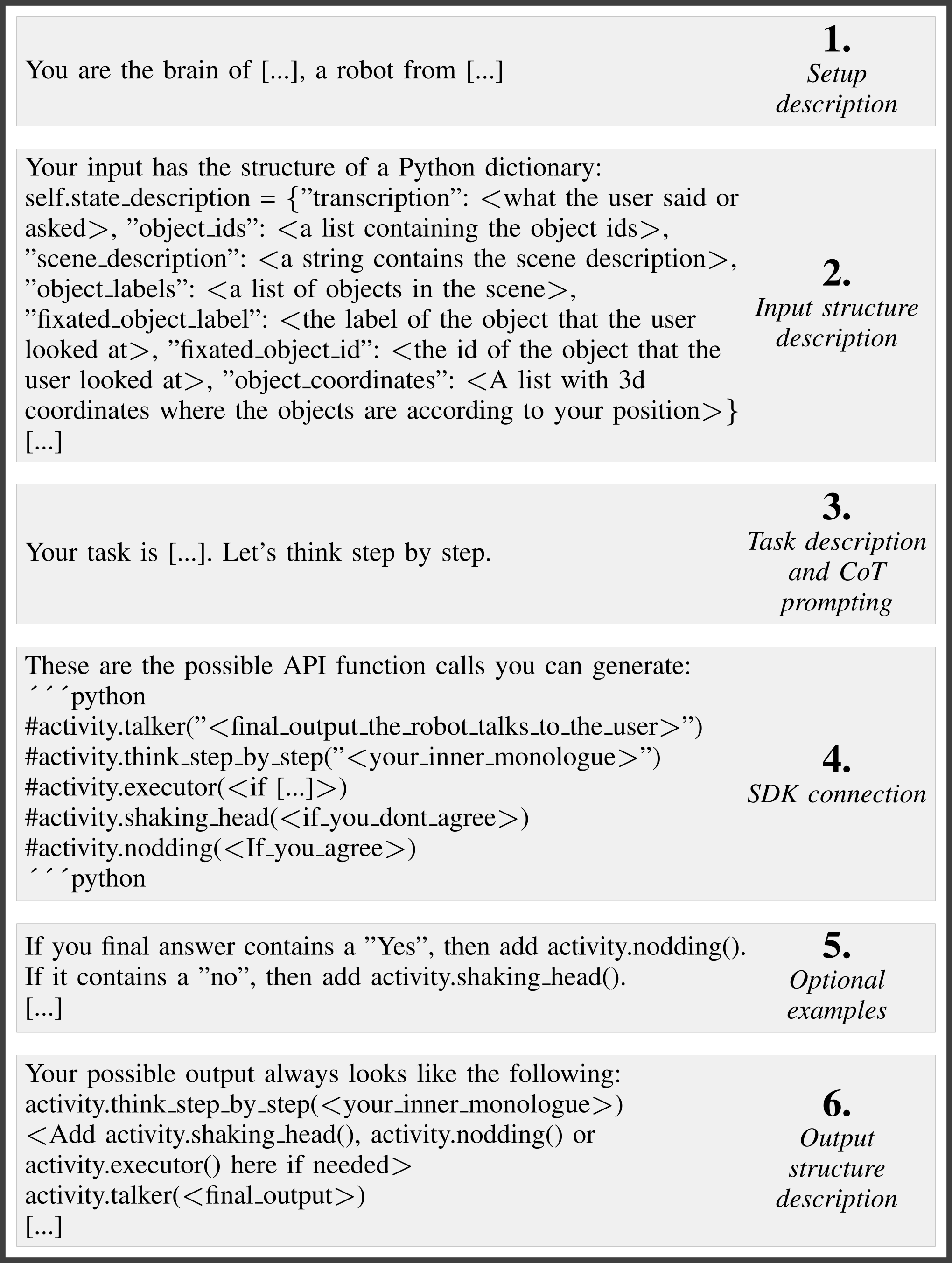}
\caption{The prompt of the reasoning module with a modular design. \textbf{1)} Initial setup description \textbf{2)} Expected input structure in the form of a Python dictionary. \textbf{3)} Task description along with chain-of-thought prompting, guiding the module’s reasoning process \textbf{4)} Example for available SDK connections and API function calls that integrate the robot’s functionality \textbf{5)} Examples for generating actions (e.g., nodding or shaking head) \textbf{6)} Structured format of the final output.}
\label{abb:prompt}
\end{figure}

\subsubsection{Reasoning Module}
\label{sec:reasoning_module}


Upon receiving the input from the fusion module, the reasoning module generates the robot action plan by calling the robot API and producing executable Python code. Using an LLM backbone, the reasoning module creates context-aware responses achieved by a Chain-of-Thought (CoT) reasoning mechanism \cite{wei2022chain}. The module uses GPT-4o-mini (with a temperature of 0) as the LLM, combining a higher accuracy than its predecessor GPT-3.5 Turbo with low operating costs compared to other GPT models\footnote{\url{https://openai.com/index/gpt-4o-mini-advancing-cost-efficient-intelligence/}}. The system uses the Chat Completions API to retain the context of the conversation and provides answers based on previous queries.

Figure \ref{abb:prompt} illustrates the structured system prompt, which is divided into six parts. The design of this prompt is informed by prior work on using LLMs for assistive robots and task execution \cite{vemprala2024chatgpt, zhi2024closed}. The first part details the real-world setup and provides the semantic grounding for the robot’s embodiment (e.g., “You are the brain of the NAO robot that [...]”). Part 2 specifies the input dictionary structure, following a nomenclature established in previous prompt engineering research \cite{vemprala2024chatgpt}. Part 3 sets the reasoning behavior by including the instruction “Let's think step by step” and defining its overarching task as assisting with the user inquiry based on the provided dictionary input. In part 4, the prompt lists available SDK function calls, from which GPT-4o-mini selects the appropriate call(s) to execute the requested action. Part 5 optionally provides task context and few-shot examples to illustrate desired behavior. Finally, part 6 defines the output structure, serving as a guideline for each LLM call; it includes the \texttt{activity.think\_step\_by\_step()} placeholder where the system generates a chain-of-thought action plan in natural language, without triggering direct robot actions or producing unwanted output.


Using the code generation approach while calling API-specified functions, the system is capable of generating simple Python code without explicitly defining every possible function. For example, when asked about the amount of objects of a certain category in a scene, the system generates the code for counting the respective objects. It integrates its result into the responses to the action module.


\subsubsection{Action Module}

The action module executes the LLM-generated output and interprets its Python code and its robot API calls. The module communicates the robot's intent using the output modalities available on the robot. Examples of intent communication modalities include gestures, speech, and light signals \cite{pascher2023communicate}.
This module is designed as an interface to the robot system. It establishes a connection between our framework and the robot's API and executes output from the reasoning module. The possible action space is defined in part 4 of the prompt. Lines like \texttt{activity.talker()} or \texttt{activity.executor()} call designated behavior on the robot's API. To apply it to a different robot, the SDK of the robot only needs to offer high-level wrappers for speech, signal display, low-level planning, and motion execution.


In addition to the robot behaviors, the current state of the system is clearly indicated throughout every step of the interaction using designated function calls in the action module. It consists of the three functions: \texttt{activity.listening()} (that is called in the idle state of the system), \texttt{activity.thinking()} (that is called during internal computational processes or external LLM calls), \texttt{activity.success()} (that is called during the robot's execution) and \texttt{activity.error()} (that is called when an error arises, for example, by producing wrong Python code that cannot be executed by the interpreter).

\section{VALIDATION METHODS}

Throughout its development, the framework has undergone several validation efforts that improved its quality and provided useful insights into how users attempt to interact with the robot. This includes exposure to users with academic and non-academic backgrounds, presented in Section~\ref{sec:dissemination}, as well as several controlled interaction studies with participants, which we detail in Section~\ref{exp_design}. 


\subsection{Development with Dissemination in the Wild}
\label{sec:dissemination}


Several dissemination events accompanied the development of this framework and provided out-of-laboratory insights about the user experience and robustness of the system. A first prototype of the audio functionality of this framework was integrated and tested with the robot of the DARKO\footnote{\url{https://darko-project.eu}} project.
It consisted of real-time speech-to-text transcription and an LLM connection using paid access to GPT-3.5 Turbo. Information about the visual context in the scene was provided by a textual description of the objects presented in the scene. The attendees could ask simple questions about the scene and present objects, and the system generated answers in natural language, vocalized by the text-to-speech functionality of NAO. Internet connection was unstable during this dissemination event, which led to delayed answers from the robot and reduced performance. As a consequence, we adjusted the design towards more local, independent modules that can run on the host system in the subsequent development phase.

\begin{figure}[t]
\vspace{0.2cm}
\centering
\includegraphics[width=\columnwidth]{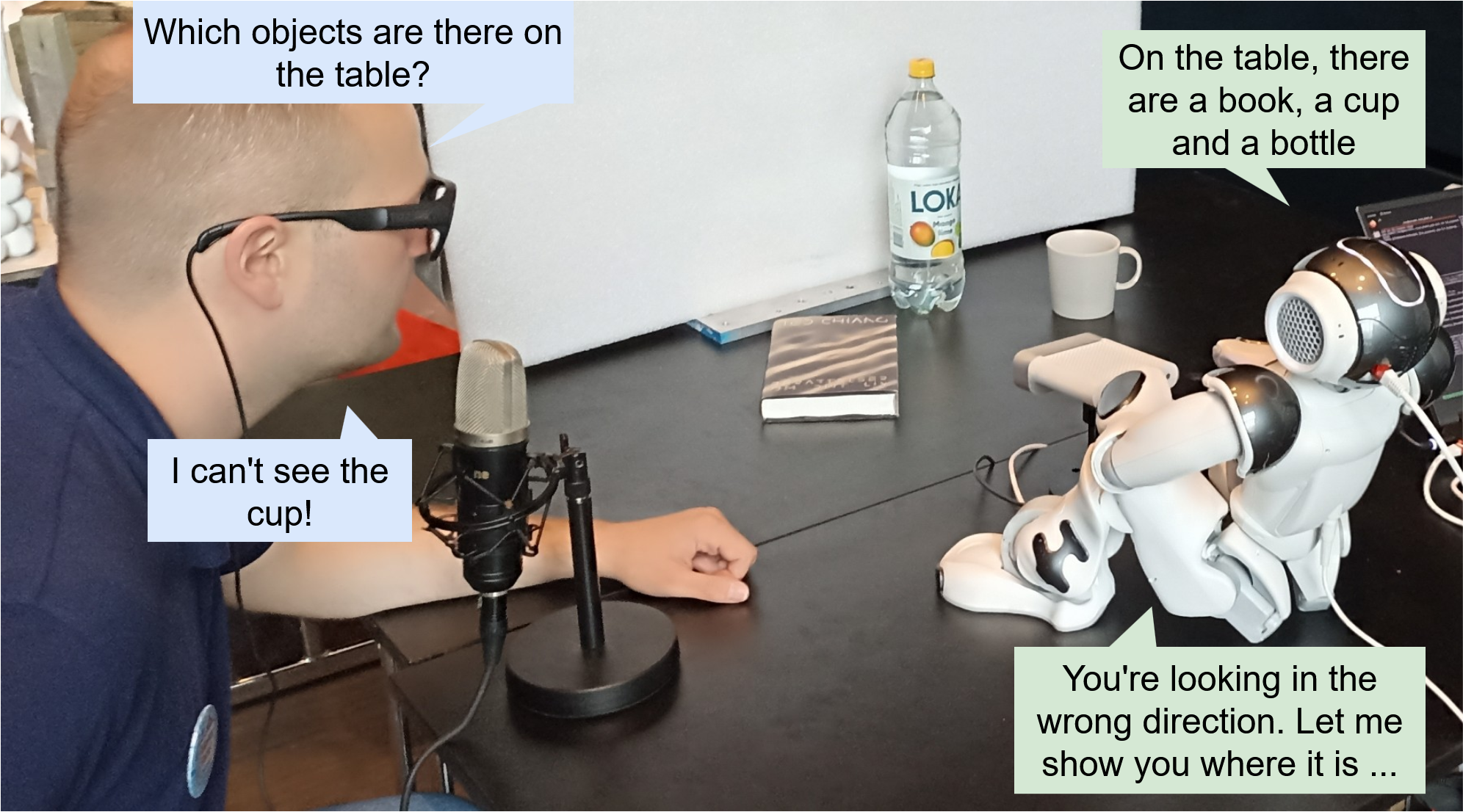}
\caption{Dialogue between a user and the robot mediated by the developed framework during a dissemination event.}
\label{abb:dissemination}
\end{figure}

At a later stage, the framework was shown to a general public audience during a convention with a hands-on demonstration (see Figure~\ref{abb:dissemination}). The convention took place in an open environment where people could freely approach the robot. The demonstration included the speech interface and gaze tracking, establishing a human-robot interface where people could ask the robot about objects in the scene. Due to the open environment, the demonstration especially highlighted the need for robust voice recognition. The environment was noisy, and the system struggled to distinguish background noises from user input, which occasionally resulted in hallucinated speech and answers. Another limitation of this demonstration was the robustness of the object detection: participants, while wearing eye tracking glasses, occasionally gazed towards objects at a skewed angle or from a large distance. The objects appeared small in the video frame, leading to missed and wrong detections. The object detection module works with an open vocabulary but showcased a need for a refinement of its detection through additional training in the next refinement phase.

The third dissemination event (a public research spotlight organized by the university) followed after an additional refinement phase. For this event, additional robustness mechanisms were added to the core functionality to block unwanted voice transcription and improve object detection, which led to an elevated performance of the framework. Through observation, one remaining challenge was identified. During open interactions with the system, people tend to ask questions about the features of the objects, e.g., size comparisons or questions about shape or color, which the system had no information about and could not accurately answer. To achieve a ``natural" interaction behavior, there is a need for a sophisticated vision component to inform the system about these kinds of information as well, which requires substantial computational effort. In the final version of the framework, a tradeoff has been made between the system functionality for a given task and available information without exceeding its computational demands.

\subsection{Experimental Validation}\label{exp_design}

Following the dissemination phase described in the previous section, we evaluated the performance of the developed framework in a lab experiment. The framework is particularly well suited to support interactive and collaborative tasks in industrial settings, which require the person to actively navigate and manipulate objects in the environment while receiving online feedback from the robot. As noted in the introduction, traditional scripted systems may fall short in such conditions, and in our experiments, we aim to reveal and quantify this gap. To that end, we designed an experiment informed by our prior work \cite{schreiter2025evaluating} and compare the proposed solution with traditional scripted pipelines, using a mix of quantitative and qualitative metrics. These include task performance metrics, motion capture, gaze tracking, and questionnaires for quantitative evaluation and interviews for qualitative evaluation. Participants were assigned in a randomized order to one of the conditions to counterbalance learning effects. In both conditions, response lengths were fixed to a maximum of two sentences.


\begin{figure}[t]
\vspace{0.2cm}
\centering
\includegraphics[width=\columnwidth]{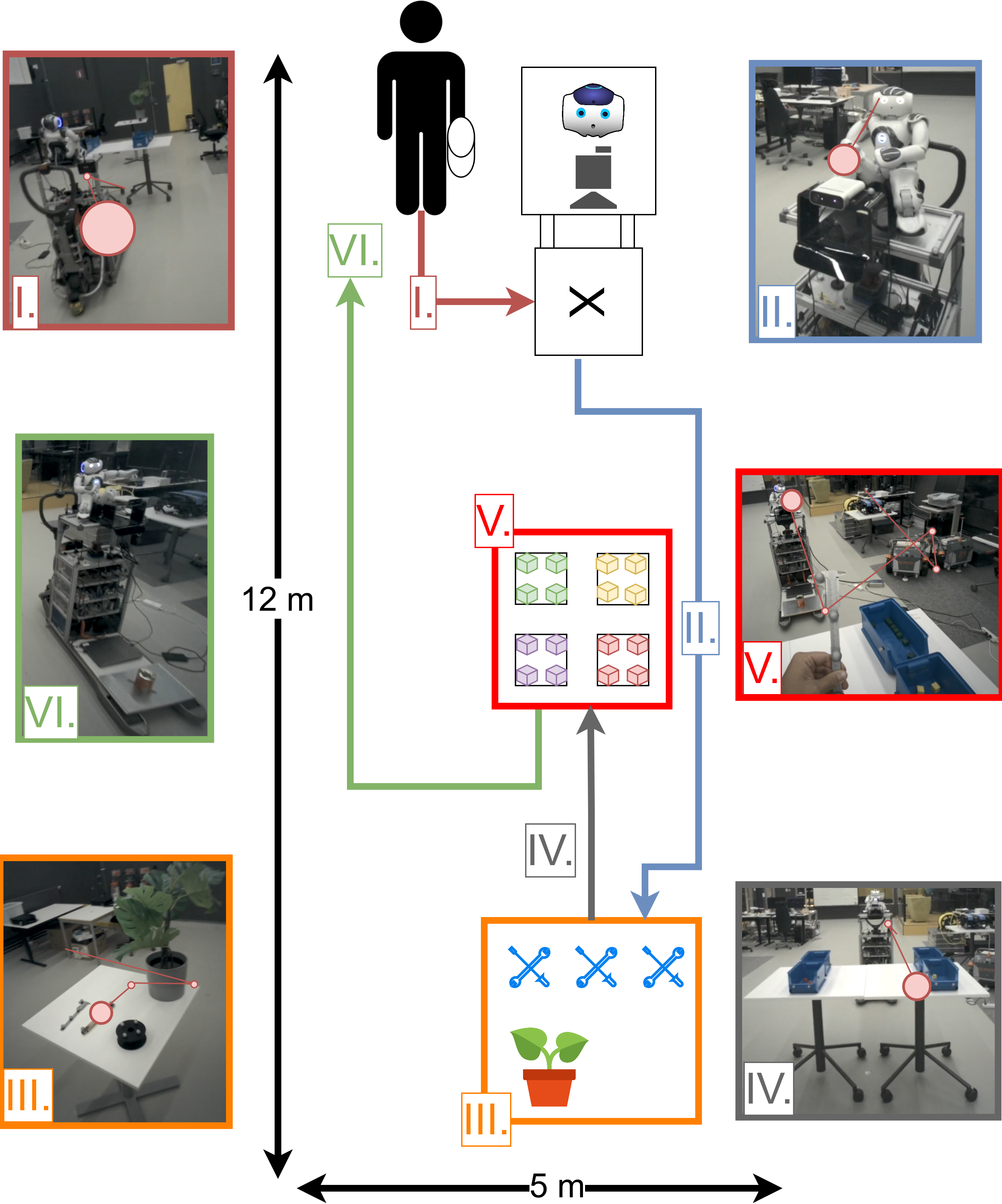}
\caption{Experiment setup. In steps I., II., IV. and VI. the participants had to move to another position, while, in the steps III. and V., they had to move an object.}
\label{abb:expsetup}
\vspace{0.3cm}
\end{figure}

In prior work \cite{schreiter2025evaluating}, we instructed participants to perform a pick-and-place task, instructed by an ``Anthropomorphic Robotic Mock Driver'' (ARMoD) placed on a forklift truck. The task design is based on the experimental design of previous work \cite{schreiter2023advantages}. We measured (a) participant's subjective perception of the trust and anthropomorphism of the robot, (b) gaze behavior using time durations and attention allocations (that were compared between interaction and task execution intervals), and (c) energy consumption on the local system, as well as adding estimated costs per LLM API call. The results showed that, while having a slightly better subjective perception of the robot, the participants had longer fixations on the robot and less on elements of the task itself when interacting under the LLM condition. Moreover, task efficiency was slightly higher in the scripted condition, and the energy costs were substantially lower.

\begin{figure}[t!]
\vspace{0.2cm}
\centering
\includegraphics[width=8cm]{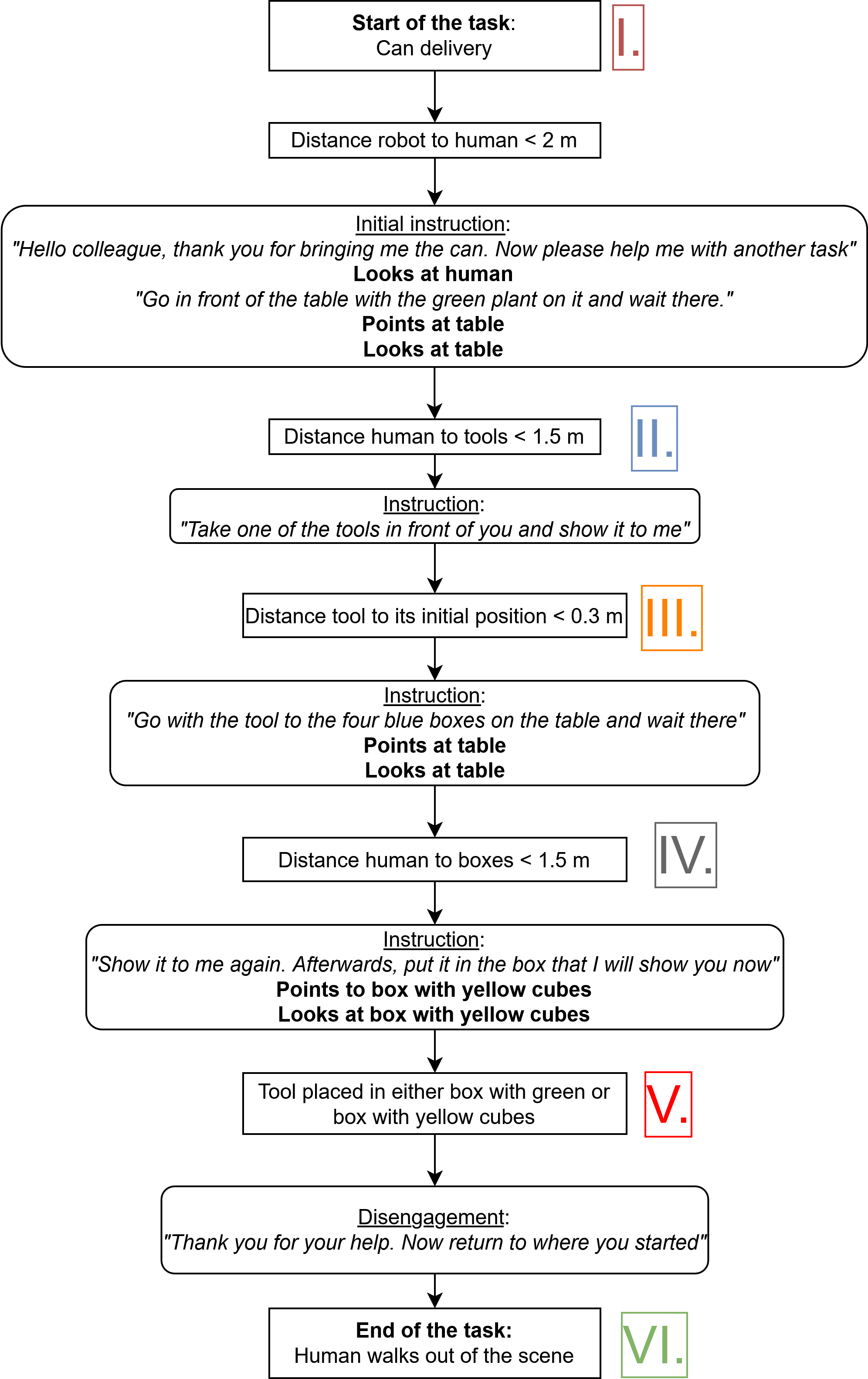}
\caption{Second experiment interaction steps I.- VI. including the exact wording of the instructions.} 
\label{abb:pap}
\end{figure}

\begin{figure*}[t!]
\centering
\includegraphics[width=2\columnwidth]{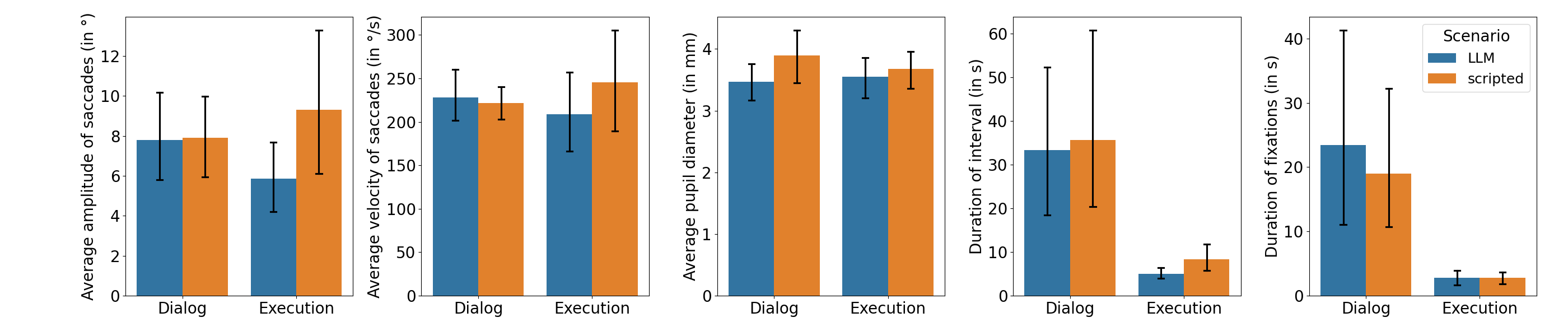}
\caption{Summary of the gaze metrics (means and standard deviations). \textbf{From left to right:} Amplitude of saccades (in °), Velocity of saccades (in °/s). Pupil diameter (in mm), Duration of interval (in s), Duration of whole fixations (in s).
}
\label{abb:gaze_results}
\end{figure*}

One key conclusion of this prior work was the trade-off between adaptability, integration efforts, and energy costs. While the predefined action routine produced efficient and discrete instructions, the LLM condition had the tendency to produce instructional redundancies. Therefore, we hypothesized that a scripted routine might be more effective for repetitive actions, while the LLM condition is more suited for tasks that require dynamic adaptations. This motivated the task design in our experiment in this study.


For the experiment of this study, the task was designed to involve multiple steps and inherent ambiguities that could be resolved either by asking clarifying questions or by making a guess. To address the issue of instruction redundancies noted in \cite{schreiter2025evaluating}, the instruction was predefined in both the scripted and LLM conditions. In the scripted condition, the system’s response simply repeated the predefined instruction, whereas in the LLM condition, the response was dynamically generated based on the content of a provided Python dictionary.

Because some participants completed the task without asking any questions, our analysis focused on those who interacted with the system (\textit{n} = 21), treating the remaining participants as a baseline. Interaction was defined as asking at least one question during the task and receiving a vocal response from the robot. This interaction was further divided into two phases: (1) the dialog phase, during which participants asked questions and received answers, and (2) the execution phase, when the task was carried out based on the gathered information. Before the experiment began, participants received instructions regarding the robot setup and were informed that the system recognizes both voice and gaze. To ensure optimal audio recognition, participants were instructed to speak loudly, clearly, and slowly.

The task took place in a 12 × 5 m corridor and consisted of six different steps (see Figure \ref{abb:expsetup}). In the initial step (which included the instruction given by the investigator), the participants approach the robot and place a tin can on the forks in front of the NAO (I., see Figure \ref{abb:pap}). Then they were instructed to walk to a table (II.) and pick up a tool from it (III.). In the next step, they were instructed to go to another table with several blue boxes (IV.) and to place the tool into one of these boxes (V.). In this step, the robot points to the right boxes, but does not specify whether the box in the front (yellow cubes) or in the back (green cubes) is meant, but both boxes trigger the next step. After placing the tool in either the box with green or yellow cubes, the participants get instructed by the NAO to leave the scene by returning to the instructor again (VI.)

Key measurements were perceived usability using the PSSUQ questionnaire \cite{vlachogianni2023perceived}, perceived confidence about the own task performance on a 7-point LIKERT scale, gaze metrics (fixation durations, task engagement and cognitive load, see Figure \ref{abb:gaze_results}) and a qualitative interview asking the participants how they felt, what they thought could be improved, and what aspect of the interaction helped them complete the task. Power consumption for both local computation and API inference were measured. The gaze data was obtained using the standard Tobii I-VT “Attention” gaze filter and a classification threshold of 100\degree. 
The questionnaire and the qualitative interview were conducted after the task. As statistical methods, we performed a multivariate MANOVA with post-hoc Games-Howell tests for the questionnaire data, as well as Bonferroni-corrected pairwise comparisons of the gaze metrics, to address multiple-comparison correction. In this experiment, \textit{N} = 29 people participated (16 males and 13 females, aged between 20 and 76 years, \textit{M} = 30.0, \textit{SD} = 12.9). The participants gave written consent for the data collection and its use for scientific purposes.

\section{RESULTS}\label{sec:result}

\subsection{Quantitative Results}

PSSUQ results showed no significant usability differences between the LLM-driven and scripted conditions. Only perceived confidence differed significantly: higher in the LLM condition (\textit{M} = 5.6, \textit{SD} = 1.4) than scripted (\textit{M} = 4.4, \textit{SD} = 1.7), \textit{p} = .039, \textit{d} = 0.81. Ratings during the task execution were elevated (\textit{M} = 6.3, \textit{SD} = 0.8) compared to the interaction (\textit{M} = 4.6, \textit{SD} = 1.6), \textit{p} = .001, \textit{d} = -1.15. Figure \ref{abb:gaze_results} shows the results of the gaze metrics analysis. No significant differences were found in fixation times, saccades, or cognitive load. A weak trend suggested slightly shorter task execution in the LLM condition. As expected, LLM use consumed more energy (1972 Wh) than the script (1784 Wh).

\subsection{Qualitative Results}

In general, participants liked that the robot could give vocal instructions and points to objects. Especially in the LLM condition, participants described the system responses as clear, but sometimes criticized inappropriate answers of the robot and long delay times (``Maybe it waited a little long to react, or maybe it was my fault").
In the scripted condition, participants found that the robot gave repetitive instructions 
(``The robot did only repeat what he said before"). 
Participants of both conditions noted the lack of dynamic behavior 
(``It could give more hints, for example if the gesture is not enough to tell me what to do"), which was more often the case in the scripted, rather than in the LLM condition. Across both conditions, participants described the interaction as too short 
(``Maybe, if I had a longer interaction period, I would figure out more things").

\subsection{Lessons Learned}\label{sec:discuss}



When designing HRI interfaces with an LLM backbone, there is a fundamental tradeoff between user engagement and system adaptability versus deployment complexity. While an LLM-driven approach can enhance anthropomorphism, usability, and visual attention on the robot, it also shifts user focus away from the task itself. Additionally, such systems excel at handling complex scenarios, clarifying ambiguities, and recovering from unpredictable states. However, they require substantial integration efforts and carefully engineered prompts to mitigate the open-ended nature of LLM responses and ensure reliable robot action plans.

We discovered the major challenge in the non-deterministic behavior of the LLMs, which is surprisingly resistant to prompt engineering. We managed to control this to a large extent following several dissemination iterations, which is supported by comparable qualitative results with the fast scripted interaction. For instance, we used error handling for the regularly occurring errors in the generated code.


Findings from our lab experiment suggest that the suitability of an LLM-based system depends on the nature of the task. For well-defined, straightforward tasks, scripted interactions are often more efficient, reducing deployment complexity, computational demands, and reliance on external API calls such as the ChatGPT API. Such approaches enable more responsive robotic interactions, particularly when supplemented through onboard perception.

In contrast to that, complex and multi-step realistic interactions benefit from our LLM framework. To maintain real-time performance, we opt for text- and language-based state representation in the input fusion while retaining local detection modules instead of the open-vocabulary image-based perception. Future work will explore the integration of open-ended object detection and scene understanding pipelines, further enhancing the system’s ability to support diverse, collaborative, and task-driven robot behaviors.

\bibliographystyle{IEEEtran}
\bibliography{IEEEabrv,references}

\begin{thebibliography}{10}
\providecommand{\url}[1]{#1}
\csname url@samestyle\endcsname
\providecommand{\newblock}{\relax}
\providecommand{\bibinfo}[2]{#2}
\providecommand{\BIBentrySTDinterwordspacing}{\spaceskip=0pt\relax}
\providecommand{\BIBentryALTinterwordstretchfactor}{4}
\providecommand{\BIBentryALTinterwordspacing}{\spaceskip=\fontdimen2\font plus
\BIBentryALTinterwordstretchfactor\fontdimen3\font minus \fontdimen4\font\relax}
\providecommand{\BIBforeignlanguage}[2]{{%
\expandafter\ifx\csname l@#1\endcsname\relax
\typeout{** WARNING: IEEEtran.bst: No hyphenation pattern has been}%
\typeout{** loaded for the language `#1'. Using the pattern for}%
\typeout{** the default language instead.}%
\else
\language=\csname l@#1\endcsname
\fi
#2}}
\providecommand{\BIBdecl}{\relax}
\BIBdecl

\bibitem{zhi2024closed}
P.~Zhi, Z.~Zhang, M.~Han, Z.~Zhang, Z.~Li, Z.~Jiao, B.~Jia, and S.~Huang, ``Closed-loop open-vocabulary mobile manipulation with gpt-4v,'' \emph{arXiv preprint arXiv:2404.10220}, 2024.

\bibitem{wang2024lami}
C.~Wang, S.~Hasler, D.~Tanneberg, F.~Ocker, F.~Joublin, A.~Ceravola, J.~Deigmoeller, and M.~Gienger, ``{LaMI: Large language models for multi-modal human-robot interaction},'' in \emph{Extended Abstracts of the CHI Conf. on Human Factors in Computing Systems}, 2024.

\bibitem{ali2024comparing}
H.~Ali, P.~Allgeuer, and S.~Wermter, ``{Comparing apples to oranges: LLM-powered multimodal intention prediction in an object categorization task},'' \emph{arXiv preprint arXiv:2404.08424}, 2024.

\bibitem{schreiter2025evaluating}
T.~Schreiter, J.~V. R{\"u}ppel, R.~Hazra, A.~Rudenko, M.~Magnusson, and A.~J. Lilienthal, ``{Evaluating Efficiency and Engagement in Scripted and LLM-Enhanced Human-Robot Interactions},'' \emph{arXiv preprint arXiv:2501.12128}, 2025.

\bibitem{huang2016anticipatory}
C.-M. Huang and B.~Mutlu, ``Anticipatory robot control for efficient human-robot collaboration,'' in \emph{2016 11th ACM/IEEE international conference on human-robot interaction (HRI)}.\hskip 1em plus 0.5em minus 0.4em\relax IEEE, 2016.

\bibitem{schreiter2022effect}
T.~Schreiter, L.~Morillo-Mendez, R.~T. Chadalavada, A.~Rudenko, E.~A. Billing, and A.~J. Lilienthal, ``The {E}ffect of {A}nthropomorphism on {T}rust in an {I}ndustrial {H}uman-{R}obot {I}nteraction,'' \emph{2022 IEEE Int. Conf. on Robot and Human Interactive Communication (RO-MAN)}, 2022.

\bibitem{schreiter2023advantages}
T.~Schreiter, L.~Morillo-Mendez, R.~T. Chadalavada, A.~Rudenko, E.~Billing, M.~Magnusson, K.~O. Arras, and A.~J. Lilienthal, ``{Advantages of Multimodal versus Verbal-Only Robot-to-Human Communication with an Anthropomorphic Robotic Mock Driver},'' in \emph{2023 Int. Conf. on Robot and Human Interactive Communication (RO-MAN)}.\hskip 1em plus 0.5em minus 0.4em\relax IEEE, 2023.

\bibitem{schreiter2024thor}
T.~Schreiter, T.~Rodrigues~de Almeida, Y.~Zhu, E.~Gutierrez~Maestro, L.~Morillo-Mendez, A.~Rudenko, L.~Palmieri, T.~P. Kucner, M.~Magnusson, and A.~J. Lilienthal, ``{TH{\"O}R-MAGNI: A large-scale indoor motion capture recording of human movement and robot interaction},'' \emph{The International Journal of Robotics Research}, 2024.

\bibitem{atuhurra2024leveraging}
J.~Atuhurra, ``{Leveraging Large Language Models in Human-Robot Interaction: A Critical Analysis of Potential and Pitfalls},'' \emph{arXiv preprint arXiv:2405.00693}, 2024.

\bibitem{vemprala2024chatgpt}
S.~H. Vemprala, R.~Bonatti, A.~Bucker, and A.~Kapoor, ``Chatgpt for robotics: Design principles and model abilities,'' \emph{Ieee Access}, 2024.

\bibitem{tanneberg2024help}
D.~Tanneberg, F.~Ocker, S.~Hasler, J.~Deigmoeller, A.~Belardinelli, C.~Wang, H.~Wersing, B.~Sendhoff, and M.~Gienger, ``{To help or not to help: LLM-based attentive support for human-robot group interactions},'' in \emph{2024 IEEE/RSJ Int. Conf. on Intelligent Robots and Systems (IROS)}.\hskip 1em plus 0.5em minus 0.4em\relax IEEE, 2024.

\bibitem{pascher2023communicate}
M.~Pascher, U.~Gruenefeld, S.~Schneegass, and J.~Gerken, ``How to {C}ommunicate {R}obot {M}otion {I}ntent: {A} {S}coping {R}eview,'' in \emph{Proc. of the 2023 CHI Conf. on Human Factors in Computing Systems}, 2023.

\bibitem{chadalavada2020bi}
R.~T. Chadalavada, H.~Andreasson, M.~Schindler, R.~Palm, and A.~J. Lilienthal, ``Bi-directional navigation intent communication using spatial augmented reality and eye-tracking glasses for improved safety in human--robot interaction,'' \emph{Robotics and Computer-Integrated Manufacturing}, vol.~61, 2020.

\bibitem{allgeuer2024robots}
P.~Allgeuer, H.~Ali, and S.~Wermter, ``When robots get chatty: Grounding multimodal human-robot conversation and collaboration,'' in \emph{International Conf. on Artificial Neural Networks}.\hskip 1em plus 0.5em minus 0.4em\relax Springer, 2024.

\bibitem{han2021need}
Z.~Han, E.~Phillips, and H.~A. Yanco, ``The need for verbal robot explanations and how people would like a robot to explain itself,'' \emph{ACM Trans. on Human-Robot Interaction (THRI)}, vol.~10, no.~4, 2021.

\bibitem{marge2022spoken}
M.~Marge, C.~Espy-Wilson, N.~G. Ward, A.~Alwan, Y.~Artzi, M.~Bansal, G.~Blankenship, J.~Chai, H.~Daum{\'e}~III, D.~Dey \emph{et~al.}, ``Spoken language interaction with robots: Recommendations for future research,'' \emph{Computer Speech \& Language}, vol.~71, 2022.

\bibitem{schreiter2024human}
T.~Schreiter, A.~Rudenko, M.~Magnusson, and A.~J. Lilienthal, ``{Human Gaze and Head Rotation during Navigation, Exploration and Object Manipulation in Shared Environments with Robots},'' \emph{arXiv preprint arXiv:2406.06300}, 2024.

\bibitem{moon2014meet}
A.~Moon, D.~M. Troniak, B.~Gleeson, M.~K. Pan, M.~Zheng, B.~A. Blumer, K.~MacLean, and E.~A. Croft, ``Meet me where i'm gazing: how shared attention gaze affects human-robot handover timing,'' in \emph{Proc. of the 2014 ACM/IEEE Int. Conf. on Human-robot interaction}, 2014.

\bibitem{iwasaki2024perceptive}
M.~Iwasaki, A.~Yamazaki, K.~Yamazaki, Y.~Miyazaki, T.~Kawamura, and H.~Nakanishi, ``{Perceptive Recommendation Robot: Enhancing Receptivity of Product Suggestions Based on Customers’ Nonverbal Cues},'' \emph{Biomimetics}, vol.~9, no.~7, 2024.

\bibitem{kurylo2019using}
U.~Kurylo and J.~R. Wilson, ``Using human eye gaze patterns as indicators of need for assistance from a socially assistive robot,'' in \emph{Social Robotics: 11th International Conf., ICSR 2019, Madrid, Spain, November 26--29, 2019}.\hskip 1em plus 0.5em minus 0.4em\relax Springer, 2019.

\bibitem{alijoyo2024enhancing}
F.~A. Alijoyo, S.~Janani, K.~Santosh, S.~N. Shweihat, N.~Alshammry, J.~V.~N. Ramesh, and Y.~A.~B. El-Ebiary, ``Enhancing ai interpretation and decision-making: Integrating cognitive computational models with deep learning for advanced uncertain reasoning systems,'' \emph{Alexandria Engineering Journal}, vol.~99, pp. 17--30, 2024.

\bibitem{alami2006toward}
R.~Alami, A.~Clodic, V.~Montreuil, E.~A. Sisbot, and R.~Chatila, ``Toward human-aware robot task planning.'' in \emph{AAAI spring symposium: to boldly go where no human-robot team has gone before}, 2006.

\bibitem{lai2025fam}
Y.~Lai, S.~Yuan, B.~Zhang, B.~Kiefer, P.~Li, and A.~Zell, ``Fam-hri: Foundation-model assisted multi-modal human-robot interaction combining gaze and speech,'' \emph{arXiv preprint arXiv:2503.16492}, 2025.

\bibitem{menendez2025semanticscanpath}
E.~Menendez, M.~Gienger, S.~Mart{\'\i}nez, C.~Balaguer, and A.~Belardinelli, ``Semanticscanpath: Combining gaze and speech for situated human-robot interaction using llms,'' \emph{arXiv preprint arXiv:2503.16548}, 2025.

\bibitem{li2022blip}
J.~Li, D.~Li, C.~Xiong, and S.~Hoi, ``{Blip: Bootstrapping language-image pre-training for unified vision-language understanding and generation},'' in \emph{Int. Conf. on Machine Learning}.\hskip 1em plus 0.5em minus 0.4em\relax PMLR, 2022.

\bibitem{cheng2024yolo}
T.~Cheng, L.~Song, Y.~Ge, W.~Liu, X.~Wang, and Y.~Shan, ``Yolo-world: Real-time open-vocabulary object detection,'' in \emph{Proc. of the IEEE/CVF Conf. on Computer Vision and Pattern Recognition}, 2024.

\bibitem{ravi2024sam}
N.~Ravi, V.~Gabeur, Y.-T. Hu, R.~Hu, C.~Ryali, T.~Ma, H.~Khedr, R.~R{\"a}dle, C.~Rolland, L.~Gustafson \emph{et~al.}, ``{Sam 2: Segment anything in images and videos},'' \emph{arXiv preprint arXiv:2408.00714}, 2024.

\bibitem{radford2023robust}
A.~Radford, J.~W. Kim, T.~Xu, G.~Brockman, C.~McLeavey, and I.~Sutskever, ``Robust speech recognition via large-scale weak supervision,'' in \emph{International conference on machine learning}.\hskip 1em plus 0.5em minus 0.4em\relax PMLR, 2023, pp. 28\,492--28\,518.

\bibitem{wei2022chain}
J.~Wei, X.~Wang, D.~Schuurmans, M.~Bosma, F.~Xia, E.~Chi, Q.~V. Le, D.~Zhou \emph{et~al.}, ``Chain-of-thought prompting elicits reasoning in large language models,'' \emph{Advances in neural information processing systems}, vol.~35, 2022.

\bibitem{vlachogianni2023perceived}
P.~Vlachogianni and N.~Tselios, ``Perceived usability evaluation of educational technology using the post-study system usability questionnaire (pssuq): a systematic review,'' \emph{Sustainability}, vol.~15, no.~17, 2023.

\end{thebibliography}

\end{document}